# *Informational non-reductionist theory of consciousness that providing maximum accuracy of reality prediction*

Vityaev E.E.
Sobolev institute of mathematics, Novosibirsk, Russia
vityaev@math.nsc.ru

*Annotation.*
The paper considers a non-reductionist theory of consciousness, which is not reducible to theories of reality and to physiological or psychological theories. Following D.I.Dubrovsky's "informational approach" to the "Mind-Brain Problem", we consider the reality through the prism of information about observed phenomena, which, in turn, is perceived by subjective reality through sensations, perceptions, feelings, etc., which, in turn, are information about the corresponding brain processes. Within this framework the following principle of the Information Theory of Consciousness (ITS) development is put forward: *the brain discovers all possible causal relations in the external world and makes all possible inferences by them.* The paper shows that ITS built on this principle: (1) also base on the information laws of the structure of external world; (2) explains the structure and functioning of the brain functional systems and cellular ensembles; (3) ensures maximum accuracy of predictions and the anticipation of reality; (4) resolves emerging contradictions and (5) is an information theory of the brain's reflection of reality.

The non-reductionist theory of consciousness is not reducible to any theory of reality and to any physiological or psychological theory.

At the Seventh International Conference on Cognitive Science, K.V.Anokhin said "The problem is not that the existing neurophysiological theories are imperfect ... The correlative approaches used in them simply cannot answer questions about the nature of mind and subjective experience ... This requires a non-reductionist fundamental theory" [3].

Max Tegmark in his book [14] also write that between the external reality "External Reality" and "Internal Reality" there should be an intermediate "Consensus Reality", describing the external reality in physical terms and at the same time is reflected in the internal reality.

Following D.I. Dubrovsky [10], we will use an "informational approach" to the description of "Internal Reality" as a *subjective reality*. Herewith, the subjective reality is the reality of an individual's conscious states – sensations, perceptions, feelings, thoughts, intentions, desires, etc. At the same time, the phenomena of subjective reality are considered as information related to the corresponding brain process as its carrier.

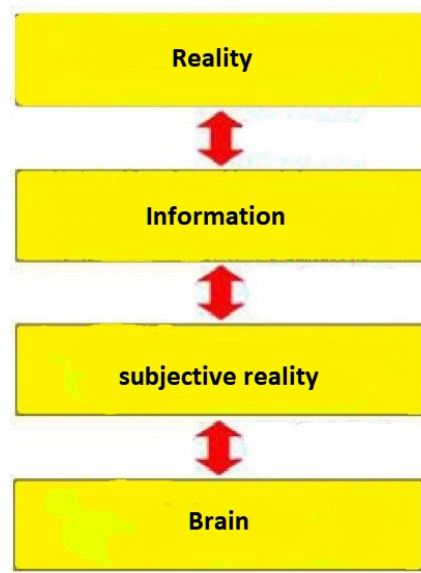

Fig. 1. Consciousness and Brain

Thus, within this framework of the information approach to the "Mind-Brain Problem", we get the following scheme fig. 1., where reality is described through information about observed phenomena, which are perceived by subjective reality through sensations, perceptions, feelings, etc., which in turn are information about the corresponding brain processes.

What is the purpose of this information? Most precisely, it is defined in the "principle of the evolution of the living world", formulated by P.K. Anokhin: "There was one universal pattern in the adaptation of organisms to external conditions, which later developed rapidly throughout the evolution of the living world: a highly rapid reflection of the slowly unfolding events of the external world" [5].

Let us reveal this principle and formulate the principles of creating an information theory of consciousness.

I. First, following the "principle of the evolution of the living world", this information theory should anticipate the events of the outside world.

However, if the external world were accidental, then anticipation of it would be impossible. But our world is well structured. If there are any laws in the information structure of reality, then it is natural to assume that in the process of evolution there were developed such neurobiological mechanisms that would use these structures to produce the most accurate reflection of reality. Therefore, the following principle is necessary.

II. The information theory of reflection should be based on the laws of the structure of the external world and describe simultaneously both the information structure of reality and neurophysiological and other mechanisms that ensure the reflection of this structure in terms of subjective reality.

The following laws of the information structure of reality will be given. The corresponding neurobiological mechanisms that use these structures were indicated by K.V. Anokhin in his report "Cognitome – hypernetwork model of the brain". These are KOGi (Cognitive Groups of Neurons), generalizing the ideas of the functional systems theory and D. Hebb's cellular ensembles [17].

We propose the following fundamental principle of the Information Theory of Consciousness (ITS), which is sufficient to explain the basic information processes:

**THE PRINCIPLE OF unlimited inference**: *The brain detects all possible causal connections in the external world and makes all possible conclusions on them.*

It turns out that this principle is sufficient to build an ITS, which:
1. Explains the structure and functioning of KOGs of functional systems and D. Hebb's cellular ensembles.
2. Based on the following information laws of the structure of the external world.
3. Provides maximum accuracy of predictions and anticipations of reality.
4. Resolves emerging contradictions.
5. It is an information theory of reflection of reality by the brain.

Let's consider the first law of the information structure of the external world – its causality. Causality is a consequence of physical determinism: "for any isolated physical system, some of its state determines all subsequent states" [12]. But consider a car accident [12]. What is the reason for it? This may be the condition of the road surface, its humidity, the position of the sun relative to the driver, reckless driving, the psychological state of the driver, brake malfunction, etc. Obviously, there is no definite reason in this case.

In the philosophy of science, causality is reduced to prediction and explanation. "Causality means predictability ... if the entire previous situation is known, the event can be predicted ... if all the facts and laws of nature related to this event are given" [12]. It is clear that to know all the facts, the number of which, as in the case of an accident, is potentially infinite and all the laws are impossible. In addition, humans and animals learn the laws of the outside world through training. Therefore, causality is reduced to prediction by inductive statistical inference, when the prediction is derived from facts and statistical laws with some probability.

In addition, causal relationships in the form of statistical laws found on real data or as a result of training face to the problem of statistical ambiguity – contradictory predictions can be derived from them [18]. To avoid this ambiguity, Hempel introduced the requirement of maximum specificity [18], informally consisting in the fact that statistical laws should include the maximum available information.

We solved the problem of statistical ambiguity and determined the Maximum Specific Causal Relationships (MSCR), for which it was proved that inductive statistical inference using them does not lead to contradictions [28,32] and thereby most accurately implement the principle of evolution of the living world by P.K.Anokhin. We have developed a special semantic probabilistic inference [28,32] that detects MSCR. In particular, it satisfies Cartwright's definition of a probabilistic causal relationship with respect to some background, which consists in the fact that each condition of the premise of causal relationship strictly increases the conditional probability of the conclusion. In addition, we have developed a formal neuron model [30] satisfying the Hebb`s rule, which implements this inference and detects MSCR causal relationships [28,32].

Thus, accurately analyzing the concept of causality, we get the informational law of reality that can be represented in ITS by detecting causal connections at the level of a neuron, ensuring its plasticity and manifesting itself, in particular, by conditional reactions. At the same time, they can be as accurate as possible through the use of MSCR conditional links.

This provides the ***first level of the maximum accuracy of predictions*** and anticipations of reality, implementing the principle of anticipatory reflection of reality.

### 3. "Natural" classification

Let's move on to the next law of the informational structure of the external world objects – the "natural" classification. The first rather detailed analysis of the "natural" classification belongs to J.S. Mill [21]. First, we will separate the "artificial" classifications from the "natural" ones: "Let's take any attribute, and if some things have it and others do not, then we can base the division of all things into two classes on it.", "But if we turn to ... the class of "animal" or "plant", ... then we will find that in this respect some classes are very different from others. ... have so many features that they cannot be ... enumerated" [21].

J.S. Mill defines the "natural" classification as follows: "Most of all, it corresponds to the goals of scientific (natural) classification when objects are combined into such groups regarding which the greatest number of general proposals can be made" [21]. Based on the concept of "natural" classification, J. S. Mill defines the concept of an "image" of a class as a certain pattern that has all the characteristics of this class.

Naturalists wrote that the creation of a "natural" classification consists in "indication" – from an infinitely large number of features it is necessary to move to a limited number of them, which would replace all other features [13]. This means that in "natural" classes, the attributes are strongly correlated, for example, if there are 128 classes and the attributes are binary, then only 7 attributes can be independent "indicator" attributes among them, since $2^7 = 128$, and other attributes can be predicted by the values of these 7 attributes. We can choose various 7-10 attributes as "indicator" and then other attributes, of which there are potentially infinitely many, can be predicted from these selected attributes. Therefore, there is an exponential (relative to the number of attributes) number of causal relationships linking the attributes of objects of "natural" classes.

Such redundancy of information, but already when perceiving objects of the external world, is confirmed in cognitive sciences when considering "natural" concepts.

### 4. "Natural" concepts in cognitive sciences

The highly correlated structure of the objects of the external world is also revealed by the theory of "natural" concepts. "Natural" classification reveals the structure of the objects of the external world, and "natural" concepts, studied in cognitive sciences, determine the perception of these "natural" objects as elements of subjective reality.

In the works of Eleanor Rosch, the following principle of categorization of "natural" categories was formulated: «Perceived World Structure … is not an unstructured total set of equiprobable co-occurring attributes. Rather, the material objects of the world are perceived to possess … *high correlational structure* … combinations of what we perceive as the attributes of real objects do not occur uniformly. Some pairs, triples, etc., are quite probable, appearing in combination … with one, sometimes another attribute; others are rare; others logically cannot or empirically do not occur» [25].

Directly perceived objects (basic objects) are information–rich bundles of observable properties that create categorization (an image in the J.S. Mill definition): «Categories can be viewed in terms of their clear cases if the perceiver places emphasis on the *correlational structure of perceived attributes* … By prototypes of categories we have generally meant the clearest cases of category membership» [24].

In further research, it was found that models based on features, similarities and prototypes are not enough to describe "natural" classes. Considering these studies, Bob Rehder put forward a theory of causal models, according to which: "people's intuitive theories about categories of objects consist of a model of the category in which both a category's features and the causal mechanisms among those features are explicitly represented" [23]. In the theory of causal models, the relation of an object to a category is no longer based on a set of signs and proximity by signs, but on the basis of the similarity of the generative causal mechanism.

Bob Rehder used Bayesian networks to represent causal knowledge [22]. However, they do not support cycles and therefore cannot model cyclic causal relationships. The formalization we propose further in the form of probabilistic formal concepts directly models cyclic causal relationships [5-8,28-29,32].

**5. The integrated information theory by G.Tononi**

The theory of integrated information by G.Tononi is also based on the highly correlated structure of the external world [20,26-27]. If the "natural" classification describes objects of the external world, and cognitive sciences describe the perception of objects of the external world, then the theory of integrated information analyzes the information processes of the brain on the perception of objects of the external world.

Integrated information is considered by G.Tononi as a property of a system of cyclic causal relationships: «Indeed, a "snapshot" of the environment conveys little information unless it is interpreted in the context of a system whose complex causal structure, over a long history, has captured some of the causal structure of the world, i.e. long-range correlations in space and time» [27].

The relationship of integrated information with reality G.Tononi describes as follows: «Cause-effect matching … measures how well the integrated conceptual structure … fits or 'matches' the cause-effect structure of its environment», «… matching should increase when a system adapts to an environment having a rich, integrated causal structure. Moreover, an increase in matching will tend to be associated with an increase in information integration and thus with an increase in consciousness» [26-27].

G.Tone defines consciousness as a primary concept that has the following phenomenological properties: composition, information, integration, exclusion [20,26-27]. We present the formulations of these properties together with our interpretation of these properties (given in parentheses) from the point of view of "natural" classification of the external world objects.

1. composition – elementary mechanisms (causal interactions) can be combined into higher-order ones ("natural" classes form a hierarchy);

2. information – only mechanisms that specify 'differences that make a difference' within a system count (only the system of "resonating" causal relationships forming the class is significant);

3. integration – only information irreducible to non-independent components counts (only the system of "resonating" causal relationships is significant, which not reducible to the information of individual components, indicating an excess of information and the perception of a highly correlated structure of a "natural" object);

4. exclusion – only maxima of integrated information count (only values of features that are maximally interconnected by causal relationships form an "image" or "prototype").

Unlike G.Tononi, we consider these properties not as internal properties of the system, but as the ability of the system to reflect the "natural" classification of the objects of the external world. Then consciousness, unlike G.Tononi, is defined not by the phenomenological properties of neural structures, but as the ability of the brain, using the integrated information of neural structures, to reflect the world represented by a hierarchical "natural" classification and the system of "natural concepts" and their causal models.

## 6. Formalization of "natural" classification, "natural" concepts and consciousness as integrated information by G.Tononi

In accordance with the Principle of unlimited inference, the brain carries out all possible conclusions on causal relationships. These causal relationships, of which there is an exponential number, in the process of perceiving "natural" objects, loop on themselves, forming a certain "resonance", which is a system with highly integrated information in the sense of G.Tononi. At the same time, "resonance" occurs if and only if these causal relationships reflect some "natural" object in which a potentially infinite set of features mutually assume each other. The resulting cycles of conclusions on causal relationships are mathematically described by "fixed points", which are characterized by the fact that further application of conclusions to the properties under consideration does not predict new properties. The set of mutually related properties obtained at a fixed point gives the "image" of the class or "prototype" of the concept and its "causal model". Therefore, the brain perceives a "natural" object not as a set of features, but as a "resonating" system of causal connections that close on themselves through the simultaneous inference of the entire set of features of the "image" or "prototype" forming a "causal model".

It can be shown that the MSCR causal relationships organized into cellular ensembles make it possible to identify objects of the external world as reliably as possible and then predict the properties of these objects as accurately as possible using this identification, since only MSPS causal relationships related to this class are used for predictions. This forms a *second*, even more accurate, from the point of view of forecasting, *level of organization of information processes*.

We propose a fundamentally new mathematical apparatus for determining integrated information, "natural" classification and "natural" concepts. Our formalization is based on a probabilistic generalization of the formal concepts analysis [8,29-32]. Formal concepts can be defined as fixed points of deterministic rules (with no exceptions) [19]. But, as J. Mill wrote: "Natural groups ... are determined by features, ... while taking into account not only the features that are certainly common to all the objects included in the group, but the whole set of those features, of which all occur in most of these objects, and the majority in all." Therefore, it is necessary to get away from deterministic rules and replace them with probabilistic ones in order to determine the features not exactly, but for the majority. Therefore, we generalize formal concepts to the probabilistic case, replacing deterministic rules with MSPS causal relationships and defining probabilistic formal concepts as fixed points of these maximally specific rules [8,29-32]. Due to the fact that the conclusions, based on the most specific causal relationships are consistent, the resulting fixed point will also be consistent and will not contain both a feature and its negation, i.e. such a definition of probabilistic formal concepts is correct.

It can be shown [9] that probabilistic formal concepts adequately formalize "natural" classification and, in moreover, the resulting "natural" classification satisfies all the requirements that naturalists imposed on "natural" classifications [9].

Let's consider an example of computer simulation of the "natural" classes, "natural" concepts and integrated information discovery for the encoded digits. Let $X(a)$ – be the set of properties of object $a$ given by some set of predicates, and let $(P_{i_1} \& ... \& P_{i_k} \Rightarrow P_{i_0}) \in MS(X)$ – be the set of MSPS of causal relationships performed for properties X, $\{P_{i_1},...,P_{i_k}\} \subset X$ then the prediction operator Pr and the fixed point can be written as follows [6,9]:

$$Pr(X) = \Phi_{Krit}(X \cup \{P_{i_0} \mid (P_{i_1} \& ... \& P_{i_k} \Rightarrow P_{i_0}) \in MS(X)\} \cup \{\neg P_{i_0} \mid (P_{i_1} \& ... \& P_{i_k} \Rightarrow \neg P_{i_0}) \in MS(X)\}),$$

where $\Phi_{Krit}(X)$ – is an operator that modifies the set of features X by adding or removing some feature, so that a certain criterion Krit of mutual consistency of causal relationships by mutual prediction of features from X is maximal [6,9]. The Krit criterion measures the informational integration of features according to the system of causal relationships differently than it is done in the G.Tononi's theory. A fixed point is reached when $Pr^{n+1}(X(a)) = Pr^n(X(a))$, for some n, where – n is a multiple application of the operator Pr. Since with each application of the operator Pr, the value of the Krit criterion increases and reaches a local maximum at a fixed point, then a fixed point, which reflect some "natural" object, has a maximum of integrated information and the "exclusion" property according to G.Tononi.

Let us encode the digits as shown in fig. 2. and form a training set, consisting of 360 shuffled digits (12 digits of fig. 2 duplicated in 30 copies without specifying where which digit is). On this set, a semantic probabilistic inference revealed 55089 MSCR causal relationships – general statements about objects that J.S. Mill spoke about.

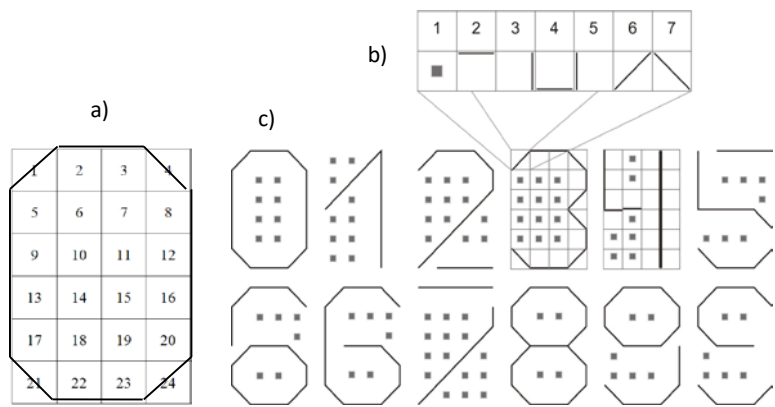

Fig. 2. Encoding of digits

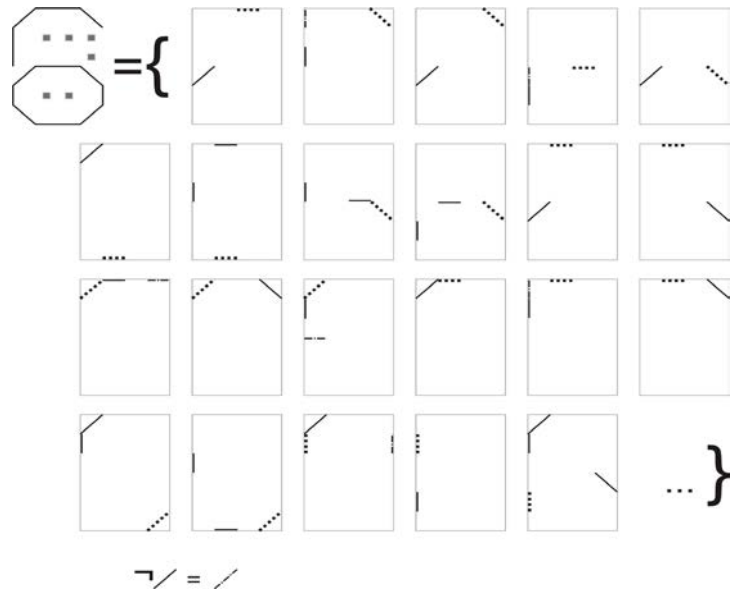

Fig. 3. The fixed point of the digit 6.

According to these causal relationships, exactly 12 fixed points were found that correspond to numbers.

An example of a fixed point for the digit 6 is shown in fig. 3. Consider what this fixed point is. Let's number the signs of the digits as shown in Fig. 2. The first pattern of figure 6 in fig. 3, represented in the first rectangle after the curly bracket, means that if there is a sign 6 in square 13 (let's denote it as 13-6), then there should be a sign 2 in square 3 (let's denote it as (3-2)). The predicted sign is indicated by a dotted line. Let's write this causal relationship as (13-6 $\Rightarrow$ 3-2). It is not difficult to verify that this causal relationship is carried out on all figures. The second causal relationship means that from the sign (9-5) and the negation of the value 5 of the first sign ¬(1-5) (the first sign should not be equal to 5), the sign (4-7) follows. Negation is indicated in the figure by a dotted line, as shown at the bottom of fig. 3. We get a causal relationship (9-5&¬(1-5) $\Rightarrow$ 4-7). The next 3 causal relationships in the first row of the digits 6 will be respectively (13-6 $\Rightarrow$ 4-7), (17-5&¬(13-5) $\Rightarrow$ 4-7), (13-6 $\Rightarrow$ 16-7).

Fig. 3 shows that the causal relationships and the signs of the number 6 form a fixed point – mutually predict each other. Note that the causal connections used in the fixed point are fulfilled on all digits, and the fixed point itself identifies only the digit 6. This illustrates the phenomenological property 2 'differences that make a difference', in which the system of causal connections perceives "realizes" an integral object. Therefore, the figures are distinguished not by causal relationships in themselves, but by their systemic relationship.

A fixed point forms a "prototype" according to Eleanor Rosch or an "image" according to J. S. Mill. The program does not know in advance which combinations of features are maximally correlated with each other.

Probabilistic formal concepts describe not only "natural" concepts, but also contexts. Contexts also have the property of maximum prediction accuracy – causal relationships found on a certain context and its causal model will more accurately predict the properties of this context.

| 1 | 2 | 3 | 4 | 5 | 6 | 7 | 8 | 9 |
|---|---|---|---|---|---|---|---|---|
| А | Б | В | Г | Д | Е | Ё | Ж | З |
| И | Й | К | Л | М | Н | О | П | Р |
| С | Т | У | Ф | Х | Ц | Ч | Ш | Щ |
| Ъ | Ы | Ь | Э | Ю | Я |   |   |   |

Fig. 4. Two contexts - Numbers and Letters

Consider the following example on fig. 4, containing both numbers and letters. You can learn only on numbers and build probabilistic formal concepts of numbers, you can learn on letters and build probabilistic concepts of letters only, and you can learn both on numbers and letters and build formal concepts of numbers and letters. In each of these cases, various MSCR causal relationships will be found, but MSCR causal relationships describing numbers and letters together will contain additional signs separating them from each other, which is obtained automatically by MSCR causal relationships. When considering (in context) both letters and numbers of MSCR causal relationships will have a higher probability, then MSCR on numbers or letters and therefore they will be triggered in the formal model of the neuron. Our formal neuron model, which detects the most specific causal connections [30], follows the well–known physiological property of neurons - more probable conditional stimuli are triggered faster in time.

### 7. Theory of functional systems

The formalization of the second type KOGs – the KOGs of functional systems, is based on the consideration of purposeful behavior, which is carried out by developing conditional (causal) links between the actions and its results. These conditional connections are sufficient for modeling functional systems and developing animats.

P.K. Anokhin wrote that "We are talking about the collateral branches of the pyramidal tract, diverting to many neurons "copies" of impulsations that go to the pyramidal tract" [4-5]. Thus, when a motor neuron sends a signal to the muscles about some action, copies of this excitation are sent, including to the projection zones, which can record the result of the action performed. Therefore, the brain detects all causal connections between actions and their results.

We show in the diagram fig. 5 that this is sufficient to explain the basic mechanisms of the functional systems of the brain formation [7,31]. Let's assume that we have no experience yet and a motivational excitement has arisen, shown by the black triangle. Then, to meet the need by trial and error, we can do some action that will be activated by some neuron, indicated by a white triangle. Simultaneously with the activation of this action, a "copy" of the excitation of this neuron will be sent to the projection zones, where there will be a neuron that will react to the result of the action received from the outside world. Since this neuron will first receive excitation from the activation of an action by a white neuron, it will form a conditional relation between the activation of an action by a white neuron and the result obtained. If now, after receiving this result that has changed the situation, we carry out some next action, also indicated by a white triangle, then we will get the following result, for which there will also be a neuron that will react to the result of this action. If, as a result, the need was satisfied and the goal is achieved, then the entire chain of active neurons and conditional connections that led to the result will be reinforced and stored in the memory. Thus, there will be an internal contour of forecasting the results achievement by the causal relationships. Then, at the next occurrence of motivational excitement, this chain of actions will be extracted from memory and will predict the achievement of the result along the inner contour even before any actions. So an action plan will be formed, which, according to the inner contour, as stated in the quote by P.K. Anokhin, activates neurons waiting for the results of actions, which will form an *acceptor of the results of actions*, studied in detail in the theory of functional systems. Thus, the formation and operation of the functional system can be explained by the formation of causal relationships between the action and its results.

In terms of MSCR causal relationships, the scheme of functional systems is as follows fig. 6. [7,31]. We consider the need as a request to the functional system to achieve the goal indicated by the predicate $PG_0$. This request enters the afferent synthesis block and, for functional systems that do not have functional subsystems, extracts causal relationships of the form $P_{i1},…,P_{im},A_{k1},…,A_{kl} => PG_0$ from memory, leading to the goal $PG_0$ achievement, where $P_{i1},…,P_{im}$ are the properties of the environment necessary to achieve the goal achievement, and $A_{k1},…,A_{kl}$ – a sequence of actions leading to the goal. At the same time, the properties of $P_{i1},…,P_{im}$ must be present in the properties of the environment $P_1,…,P_n$ entering the afferent synthesis block. For hierarchically organized functional systems, this query extracts causal relationships of a more complex type $P_{i1},…,P_{im},PG_{j1},…,PG_{jn},A_{k1},…,A_{kl} => PG_0$ from the MSCR memory, including requests to achieve the sub-goals $PG_{j1},…,PG_{jn}$.

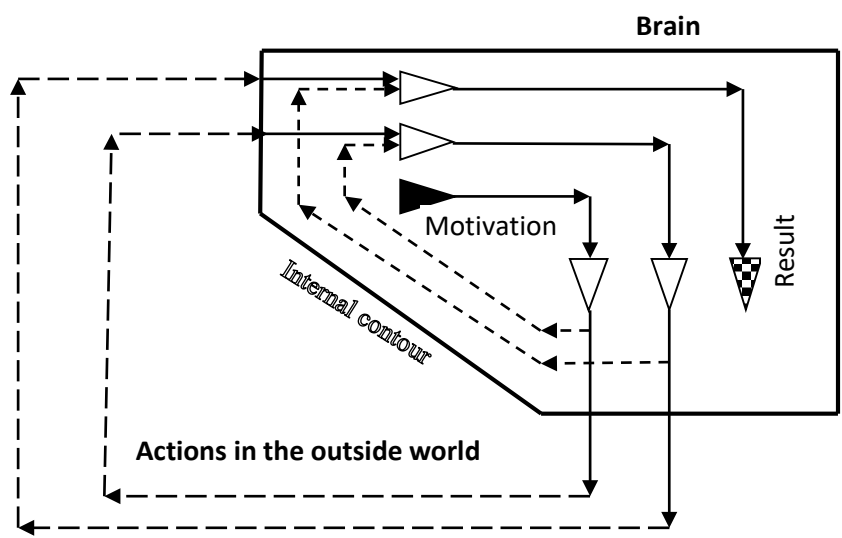

Fig. 5. Formation of MSCR of conditional connections between actions actions and results

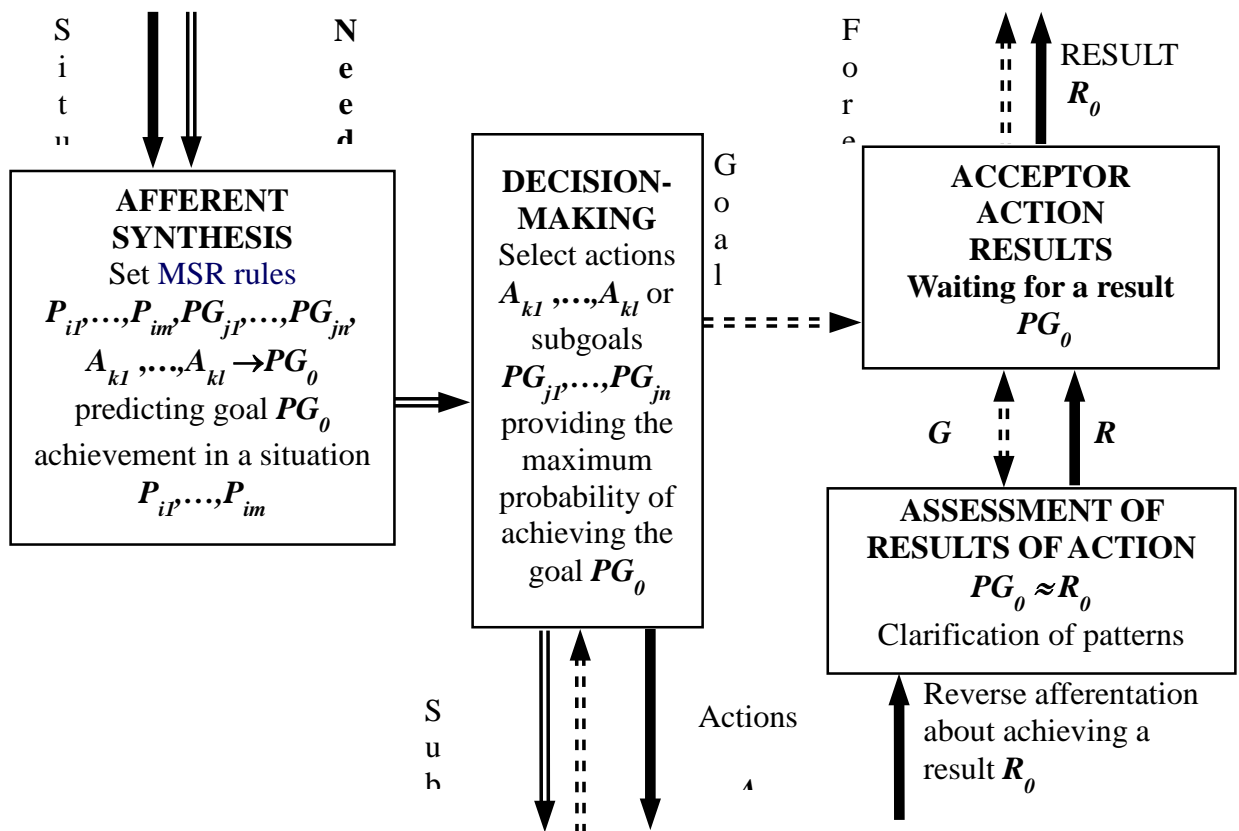

Fig. 6. Functional system diagram

Then the extracted rules are sent to the decision-making block, where forecast of the goal achievement is made for each rule and a probability estimation of the goal achievement is calculated. Prediction according to the rules, where only actions are performed, is carried out according to the probability of the rule itself. The forecast according to the rules which requests sub-goals is carried out by sending these requests to functional subsystems, decision-making in them and receiving from them the probability estimations of the corresponding sub-goals achievement. The resulting probability of the forecast is calculated by the product of the probability of the rule on the probability of achieving of its sub-goals. After that, a decision is made by the rule selecting that has the maximum probability estimation of the goal achievement.

Then an action plan is formed, including all the actions included in the rule and all the actions that are in the functional subsystems. Simultaneously with the action plan, the acceptor of the actions result is formed, including the expectation of all predicted sub-results in functional subsystems, as well as in the functional system itself. After that, the action plan begins to be implemented, and the expected results are compared with the results obtained.

If all the sub-results and the final result are achieved and coincide with the expected results, then the rule itself and all the rules of the functional subsystems that were selected in the decision-making process are reinforced and their probability increases. If the result is not achieved in some subsystem, then the corresponding rule, selected by decision-making block, of this functional subsystem is penalized. Then there is a tentative research reaction that revises the decision. This model has been successfully used to model animates [11,16,31].

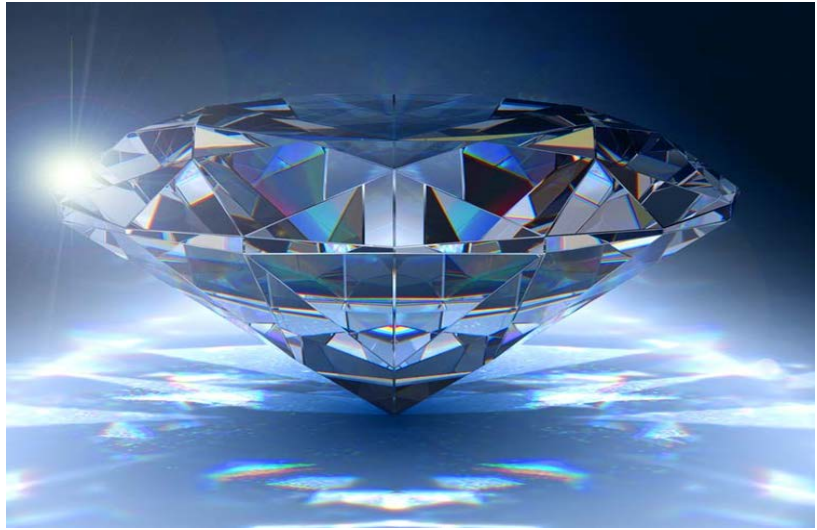

Fig. 7. Multi-faceted reality.

## 8. Consciousness as the tool for resolving contradictions

Let us consider the *third level of prediction accuracy* provided by the information theory of consciousness – consciousness as a mechanism for resolving contradictions.

The world is multifaceted like a diamond fig. 7 and there is no single consistent description of it, and the function of consciousness is to choose the appropriate context correctly, within which you can get the most accurate prediction. In science, such contexts are paradigms that form a certain view, point of view and the corresponding system of concepts of a particular theory. These paradigms, as a rule, are not compatible with each other.

This point of view on consciousness is also expressed by V.M. Allakhverdov [2]. In his work [1] he writes: "Consciousness, faced with contradictory information, tries to remove this information from the surface of consciousness or modify it so that the contradiction disappears or ceases to be perceived as a contradiction." In this work, he cites 7 cases of resolving contradictions by consciousness. All these cases are explained by the properties or interaction of probabilistic formal concepts that define the concepts or contexts in question. Let 's consider two of them for brevity:

1. **Case 1**. The easiest way to get rid of contradiction or ambiguity is to choose one interpretation for awareness, and not to realize all the others (incompatible with it) (negatively choose).

   *Example*. The phenomenon of binocular competition, when different stimuli are simultaneously presented to the different eyes to the subject. If two images are presented, one of which is more likely or familiar, the subjects mostly see only it.

   *Explanation*. A probabilistic formal concept mutually predicts the properties included in the concept, as well as the negation of other properties that should not be in it, and thereby inhibits alternatives.

2. **Case 2**. When realizing the different sides of the contradiction, an attempt is made to find a way of explanation — the connection of different sides into a consistent whole.

*Example*. In the conditions of binocular competition, if you present a red circle on one eye and a black triangle on the other, the subject will see a black triangle on a red background.

*Explanation*. If the perceived features do not contradict each other and do not inhibit each other, then they can form a combined probabilistic formal concept and be perceived accordingly.

In all cases, it is possible to correctly choose the most appropriate probabilistic formal concept or context to resolve the contradiction and obtain the most accurate prediction in accordance with the causal relationships of the chosen concept or context.